%% file: ms.tex
\documentclass[conference]{IEEEtran}
\IEEEoverridecommandlockouts
\usepackage{cite}
\usepackage{amsmath,amssymb,amsfonts}
\usepackage{algorithmic}
\usepackage{graphicx}
\usepackage{textcomp}
\usepackage{xcolor}
\usepackage{caption}
\usepackage{subcaption}
\def\BibTeX{{\rm B\kern-.05em{\sc i\kern-.025em b}\kern-.08em
    T\kern-.1667em\lower.7ex\hbox{E}\kern-.125emX}}
\def\smallspace{\hspace{.1em plus .1em}}
\begin{document}

\title{Towards Tenodesis-Modulated Control of an Assistive Hand Exoskeleton for SCI\\

\thanks{This work was supported by the National Institute of Neurological Disorders and Stroke under grant R01NS115652.}%
\thanks{$^{1}$Department of Mechanical Engineering, Columbia University, New York, NY 10027, USA.%
}
\thanks{$^{2}$Department of Rehabilitation and Regenerative Medicine, Columbia University, New York, NY 10032, USA.%
}
\thanks{$\dagger$ Joint lead authors. \texttt{\{jbp2157, ad3805\}@columbia.edu}}
}

\author{Joaquin Palacios$^{\dagger,1}$, Alexandra Deli-Ivanov$^{\dagger,1}$, Ava Chen$^{1}$, Lauren Winterbottom$^{2}$,\\ Dawn M. Nilsen$^{2}$, Joel Stein$^{2}$, and Matei Ciocarlie$^{1}$
}

\maketitle

\input{Sections/Introduction}
\input{Sections/Methods}
\input{Sections/Results}
\vspace{-1.5mm}
\input{Sections/Conclusion}

\end{document}

%% file: Sections/Introduction.tex
\section{Introduction}
\vspace*{-3pt}
A Spinal Cord Injury (SCI) can have life-altering consequences, and with an estimated 18,000 yearly cases in the US, the societal impact cannot be overstated \cite{SCI cases US}. SCI often results in partial or complete sensorimotor loss in the arms and body, leading to limited independence. As such, restoration of hand function is one of the highest priorities for SCI populations~\cite{SCI Priorities Review}.

Many individuals with C6-C7 SCI have preserved wrist mobility and use tenodesis to grasp. Tenodesis can achieve some degree of lateral pinch and grasp by exploiting the passive finger flexion that occurs when the wrist is extended. However, the grasping forces generated often fall short of what is required for activities of daily living (ADLs), even when wrist torque is amplified by unmotorized devices such as Wrist-Driven Flexor Hinge Orthotics (WDFHOs) \cite{Tenodesis and WDFHO Forces}. The WDFHO design still remains popular as it provides a familiar mechanism for grasp force modulation with wrist motion, despite being restricted in force output due to the wrist's limited range of motion \cite{Tenodesis and WDFHO Forces}.

Robotic assistive devices for SCI have been more successful at generating the forces required for ADLs, but independent user operation poses a challenge \cite{Review Paper}. Tenodesis can be leveraged as an intuitive and familiar control method, using the wrist as a control input while decoupling it from force generation. Many assistive devices, however, cannot exploit this due to obstructions to wrist mobility \cite{Review Paper}. While a tenodesis-inspired wrist user control has been seen with robotic devices \cite{Hybrid Orthosis, Exo-Glove, Tenodesis-Grasp-Emulator}, testing with SCI users remains underexplored. Moreover, grasping is often treated as a binary (either open or closed), not taking into account variations in grasp force for different objects. Integrating the force modulation of WDFHOs could potentially improve end-users' intuitive control, comfort, and safety utilizing assistive devices. 

\begin{figure}[t]
\centering
\includegraphics[trim=0cm 0cm 0cm 7cm,clip=true,width=\columnwidth]{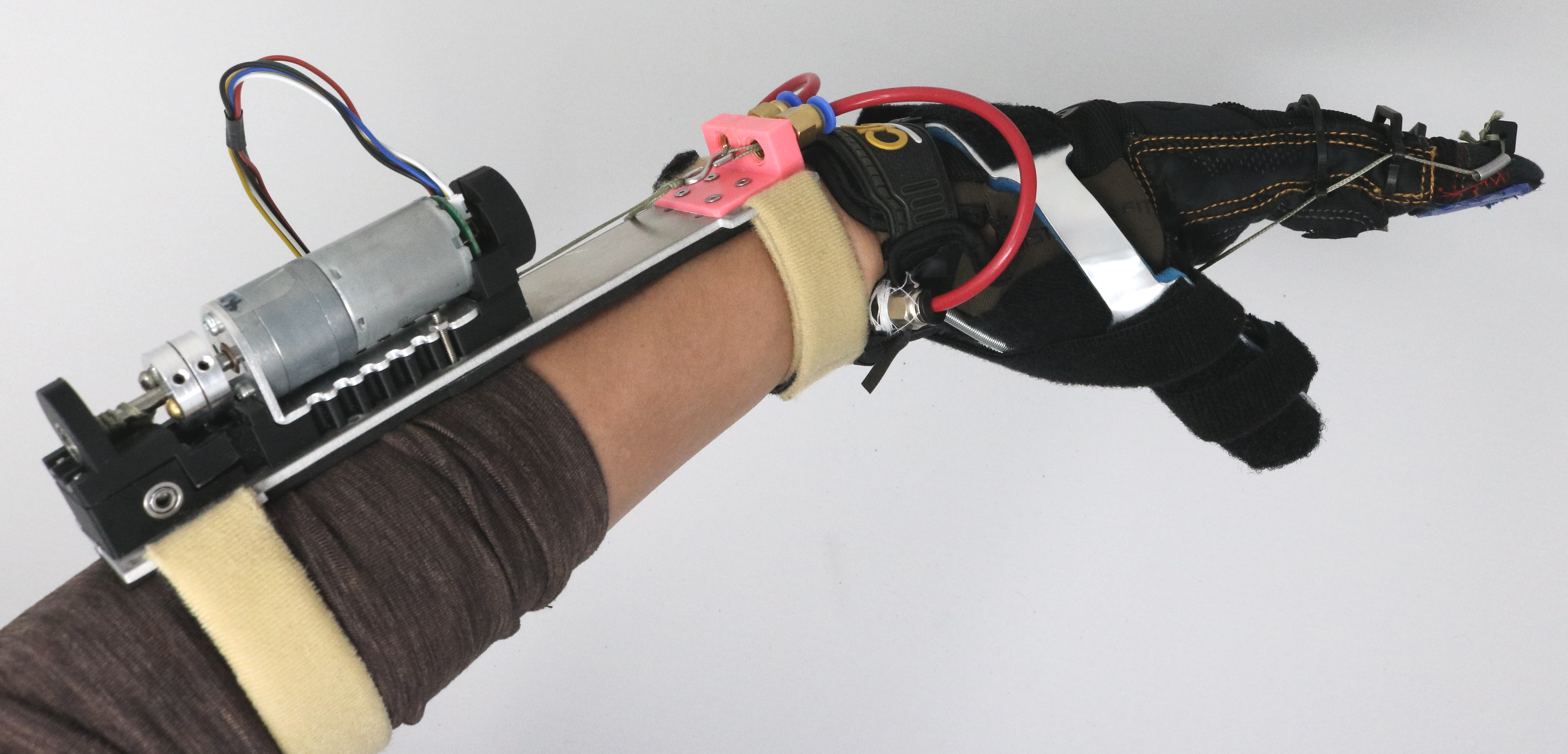}
\caption{MyHand-SCI assists finger flexion for grasping without encumbering wrist motion.}
\label{Glamour Shot}
\centering
\vspace*{-12pt}
\end{figure}

In this work, we present a prototype of a robotic assistive orthosis capable of implementing tenodesis user control. The underactuated device provides active grasping assistance while preserving free wrist mobility through the use of Bowden cables. This device enables force modulation during grasping, which was effectively leveraged by a participant with C6 SCI to demonstrate improved grasping abilities using the orthosis. 

%% file: Sections/Methods.tex
\section{MyHand-SCI Assistive Device}
MyHand-SCI is intended to assist pinch and power grasps. These grasps are versatile in terms of the sizes and types of objects a user can grab and utilize for ADLs. Comfort and ease of use were important design criteria, which motivated our under-actuation scheme where a single motor closes multiple fingers in a synergistic grasping motion. Finally, a primary objective was preserving free wrist mobility to enable emulating a tenodesis grasp.

\subsection{Device Design} 
\begin{figure}[t]
    \centering
    \includegraphics[width=\columnwidth]{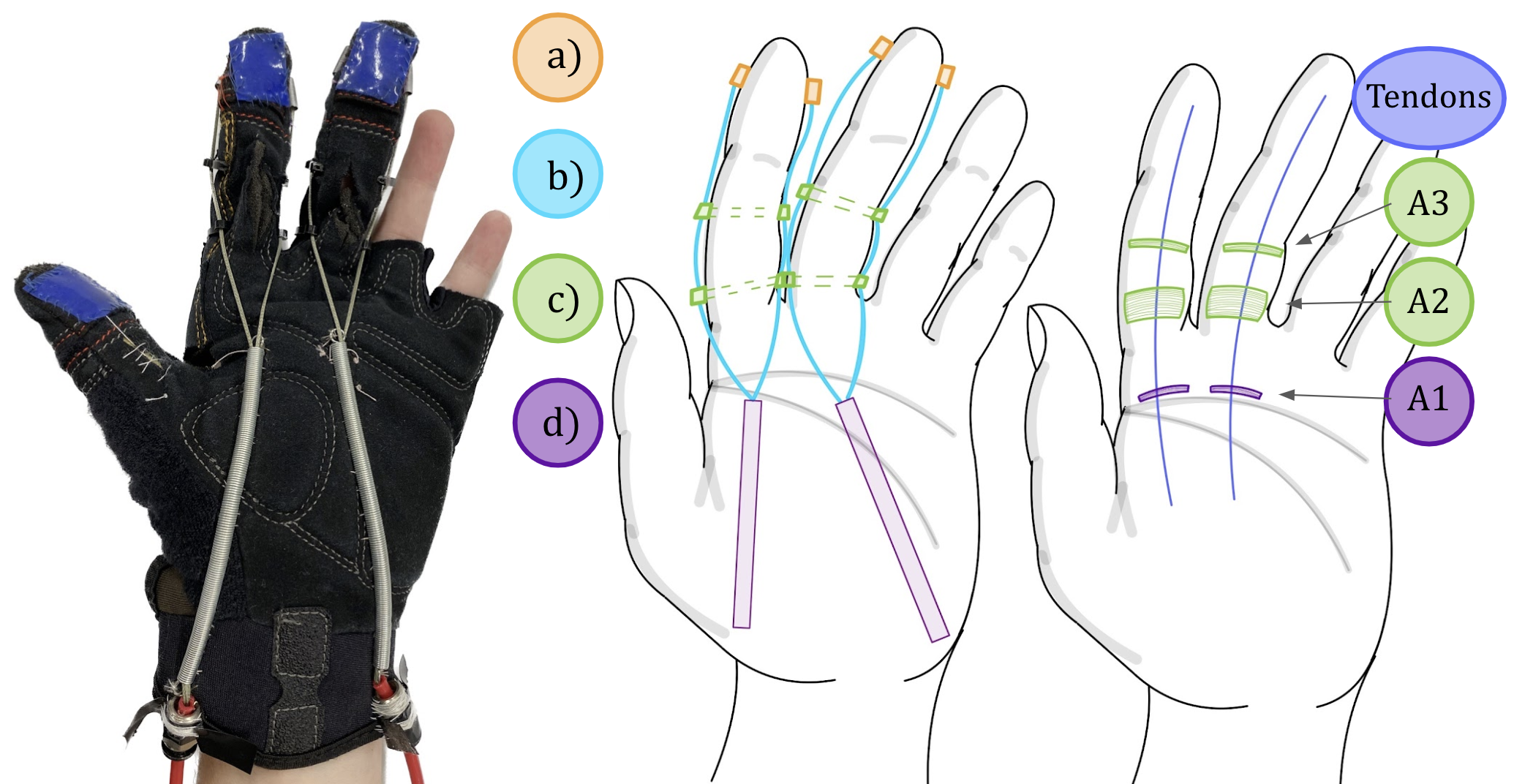}
\vspace{-5mm}
    \caption{Bioinspired tendon routing. Left: Photo of device. Middle: Device diagram shows a) Termination point, b) Exotendons, c) Guide rings, d) Bowden cable. Right: Simplified diagram with labeled anatomical tendon pulleys.}
    \label{device_diagram}
\vspace{5mm}
    \includegraphics[trim=0cm 0cm 0cm 8mm,clip=true,width=\columnwidth]{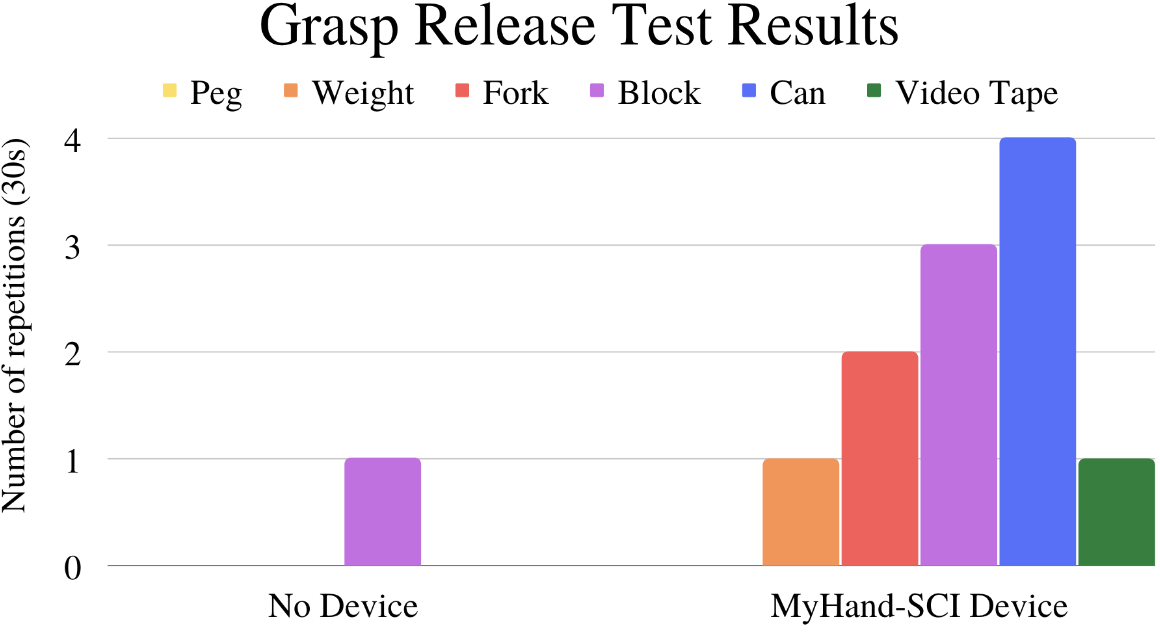}
    \vspace{-6mm}
    \caption{Grasp and Release Test scores with one participant with C6 tetraplegia show performance for each test object.}
    \vspace{1mm}
    \label{fig: GRT Results}
\vspace{-2mm}
\end{figure}

The MyHand-SCI is a fabric-based, soft robotic, exotendon-driven assistive orthosis (Figure \ref{Glamour Shot}). This prototype actuates digits 2 and 3 via exotendons routed from each finger to a motor attached to a dorsal forearm splint (Figure \ref{Glamour Shot}). The splint terminates before the wrist joint to enable free mobility, which is critical to tenodesis motion. Starting from the motor, the tendons route through Bowden cables, around the wrist, through the palm, and up the sides of the fingers. Bowden cables provide a flexible, low-profile, and low-friction pathway through the hand that protects the tendons and minimizes interference with the user's ability to grasp objects or extend their wrists. Our bioinspired routing follows the anatomy of the finger by configuring guide rings to roughly mimic the location of the anatomical A1, A2, and A3 tendon ligaments (Figure \ref{device_diagram}). These guide rings act as exotendon pulleys that encourage natural finger flexion. Termination at the fingertip improves leverage, giving strength to our desired grasps. The thumb is splinted in opposition to support power and pinch grasps. The device is currently operated through a two-button control, which allows the user to modulate the degree of finger flexion and the grasping force applied. The overall device weighs 295 grams (excluding the tabletop controller and power supply). 

%% file: Sections/Results.tex
\section{Research Participant Testing}

Our device prototype was tested on one individual with C6 tetraplegia using the Grasp and Release Test (GRT). The GRT is a standardized test that assesses the ability of an individual with C5 or C6 tetraplegia to manipulate a set of objects varying in size and weight.
Human subject testing was conducted at the Columbia University Irving Medical Center in accordance with the protocol \mbox{(IRB-AAAU2339)} approved by the Columbia University Institutional Review Board. The assistive device was operated by a researcher who responded to verbal cues from the participant to open or close the hand. The operator had the ability to modulate the degree of finger flexion and the force exerted to ensure participant comfort.

\begin{figure}[t]
  \begin{center}
  \includegraphics[scale =0.3]{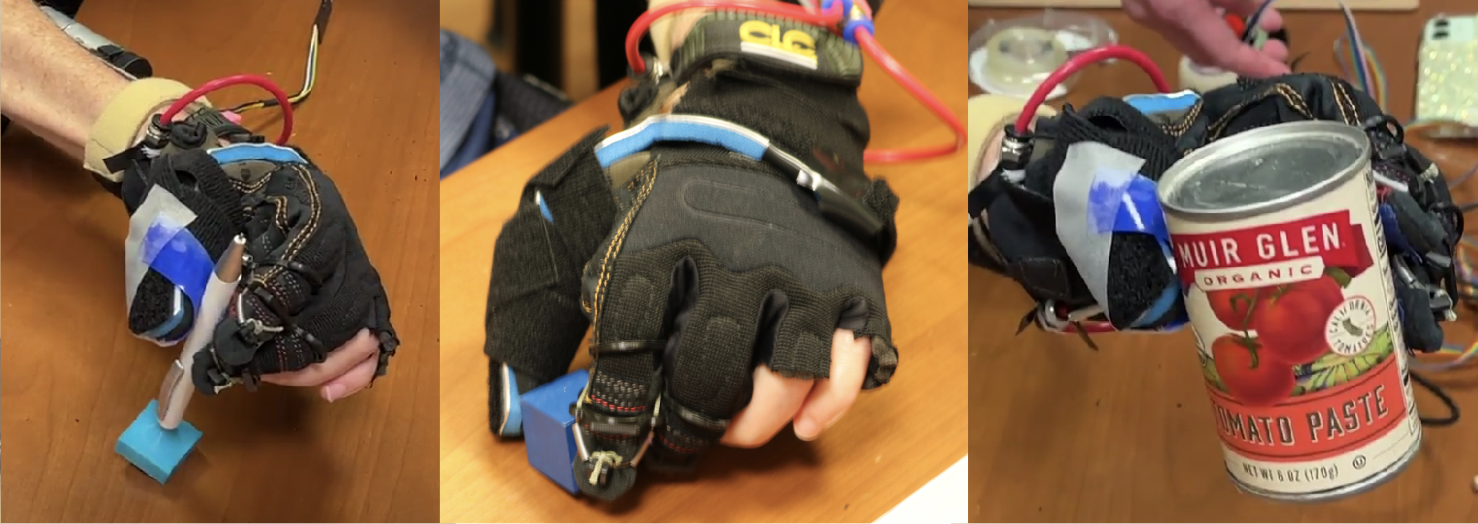}
  \caption{The participant using the device to interact with objects.}
  \label{fig: GRT objects}
  \end{center}
  \vspace*{-22pt}
\end{figure}

Within one session, the participant obtained a score of 11 with the assistive device compared to 1 without it. The scores reflect the cumulative amount of times the participant successfully grasped and released the items, given a 30-second interval for each object. Figure \ref{fig: GRT Results} shows specific performance per item and Figure \ref{fig: GRT objects} provides images of the participant grasping several of the objects in the assessment. The participant responded well to the device, showing no signs of discomfort or adverse effects, and was able to determine and modulate (through verbal cues) how much force he needed from the device. The participant's wrist mobility and range of motion were undisturbed by the device. 

\vspace{-1.5mm}
Our GRT assessment shows an improvement in the participant's ability to handle everyday objects. This validates our mechanical design and provides initial support for the orthosis' effectiveness as an assistive device. The participant was able to effectively communicate with the researcher while using the device, which helped cater to various preferences and mitigate fatigue or discomfort.
Moreover, grasping modulation proved to be an advantageous feature worthy of further exploration. With a functioning device, we are in a position to incorporate and test features that allow a user to make use of a tenodesis grasp to control their finger movements.
\vspace{-0.15mm}

%% file: Sections/Conclusion.tex
\section{Future Work}

\vspace{-1.mm}
Given the promising results of our first session with a participant, our initial prototype appears to provide effective grasping assistance in a lightweight and comfortable form factor. The next step is to implement our proposed tenodesis control mode. One potential implementation would be integrating sensors to extract the angle of wrist extension and map this proportionally to the extent of finger flexion and force applied during grasping. Additional user control features could be integrated, including a pattern recognition modality. For instance, two successive quick wrist extensions could activate a grasp-locking mode, where a stable grasping configuration is maintained regardless of wrist motion. This would allow users to grasp objects without continually exerting their wrists, reducing fatigue and improving usability. We envision a first round of testing with healthy subjects to ensure safety and assess how intuitive this control method is, followed by sessions with individuals with SCI examining the same questions.